\documentclass[runningheads]{llncs}
\usepackage{graphicx}
\usepackage{cite} 
\usepackage{times}
\usepackage{epsfig}
\usepackage{amsmath}
\usepackage{amssymb}
\usepackage{mwe}
\usepackage{acro}
\usepackage{amssymb}
\usepackage{xcolor,colortbl}
\usepackage{tabularx}
\usepackage{relsize}
\usepackage{pifont}
\usepackage{booktabs} 
\usepackage{multirow}
\usepackage{multicol}
\usepackage{adjustbox}
\usepackage{float}
\usepackage[colorlinks,
            linkcolor=red,  
            anchorcolor=blue, 
            citecolor=green,   
            ]{hyperref}
\usepackage[flushleft]{threeparttable}

\begin{document}
\title{Cross-scale Attention Guided Multi-instance Learning for Crohn's Disease Diagnosis with Pathological Images}
%
%
\author{Ruining Deng\inst{1} \and
Can Cui\inst{1} \and
Lucas W. Remedios \inst{1}\and 
Shunxing Bao \inst{1} \and
R. Michael Womick \inst{2} \and
Sophie Chiron \inst{3} \and 
Jia Li \inst{3} \and
Joseph T. Roland \inst{3}  \and
Ken S. Lau \inst{1}\and
Qi Liu \inst{3}\and
Keith T. Wilson \inst{3,4} \and 
Yaohong Wang \inst{3} \and  
Lori A. Coburn \inst{3,4} \and 
Bennett A. Landman \inst{1}  \and 
Yuankai Huo \inst{1}}

%

\institute{
1. Vanderbilt University, Nashville TN 37215, USA, \\
2. The University of North Carolina at Chapel Hill, Chapel Hill,
NC, 27514, USA, \\
3. Vanderbilt University Medical Center, Nashville TN 37232, USA, \\
4. Veterans Affairs Tennessee Valley Healthcare System, Nashville, TN, 37212, USA}


\maketitle              
\begin{abstract}

Multi-instance learning (MIL) is widely used in the computer-aided interpretation of pathological Whole Slide Images (WSIs) to solve the lack of pixel-wise or patch-wise annotations. Often, this approach directly applies ``natural image driven" MIL algorithms which overlook the multi-scale (i.e. pyramidal) nature of WSIs. Off-the-shelf MIL algorithms are typically deployed on a single-scale of WSIs (e.g., 20$\times$ magnification), while human pathologists usually aggregate the global and local patterns in a multi-scale manner (e.g., by zooming in and out between different magnifications). In this study, we propose a novel cross-scale attention mechanism to explicitly aggregate inter-scale interactions into a single MIL network for Crohn's Disease (CD), which is a form of inflammatory bowel disease. The contribution of this paper is two-fold: (1) a cross-scale attention mechanism is proposed to aggregate features from different resolutions with multi-scale interaction; and (2) differential multi-scale attention visualizations are generated to localize explainable lesion patterns. By training $\sim$250,000 H\&E-stained Ascending Colon (AC) patches from 20 CD patient and 30 healthy control samples at different scales, our approach achieved a superior Area under the Curve (AUC) score of 0.8924 compared with baseline models. The official implementation is publicly available at \url{https://github.com/hrlblab/CS-MIL}.

\end{abstract}

\section{Introduction}

\begin{figure}
\begin{center}
\includegraphics[width=0.9\textwidth]{{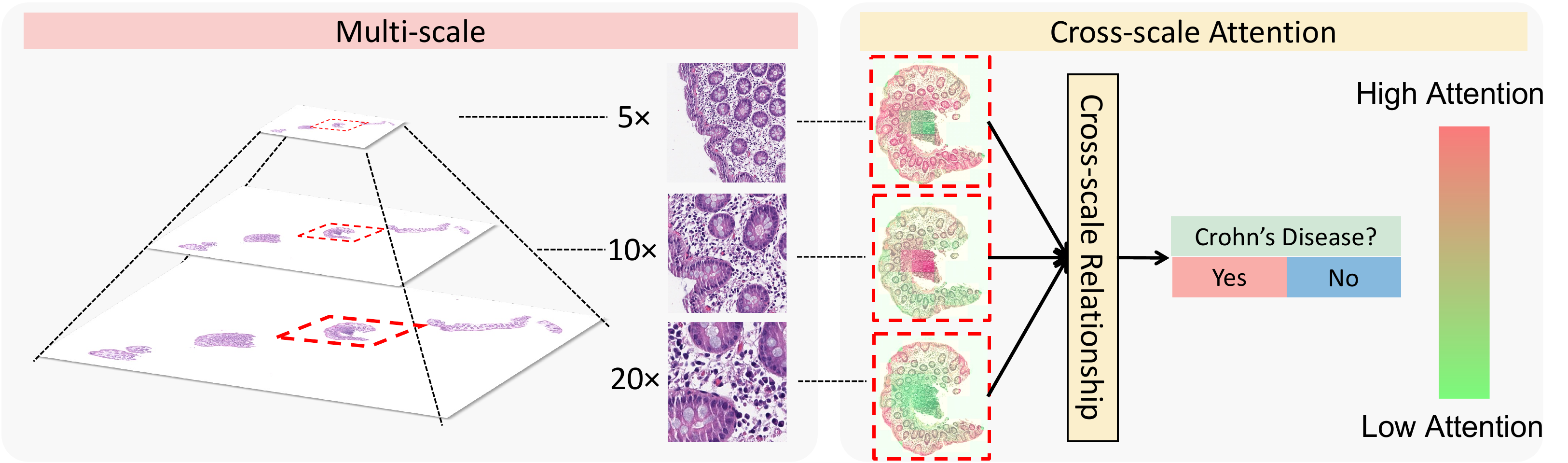}}
\end{center}
\caption{\textbf{Multi-scale awareness.} Human pathologists typically aggregate the global and local patterns in a multi-scale manner. However, previous work failed to be aware of cross-scale relationship at different resolutions. Our method demonstrates the importance-of-regions with cross-scale attention maps, and aggregate the multi-scale patterns with differential attention scores for CD diagnosis.} 
\label{fig1:Problem}
\end{figure}

Digital pathology is relied upon heavily by clinicians to accurately diagnose Crohn’s Disease (CD)~\cite{gubatan2021artificial,yeshi2020revisiting}. Pathologists carefully examine biopsies at multiple scales through microscopes to examine morphological patterns ~\cite{bejnordi2017diagnostic}, which is a laborious task. With the rapid development of whole slide imaging (WSI) and deep learning methods, computer-assisted CD clinical predicion and exploration~\cite{kraszewski2021machine,con2021deep,kiyokawa2022deep,syed2020potential} are increasingly promising endeavors. However, annotating images pixel- or patch-wise is computationally expensive for a standard supervised learning system~\cite{hou2016patch,mousavi2015automated,maksoud2020sos,dimitriou2019deep}. To achieve accurate diagnoses from weakly annotated images (e.g., patient-wise diagnosis), multi-instance Learning (MIL) -- a widely used weakly supervised learning paradigm -- has been applied to digital pathology ~\cite{wang2019rmdl,skrede2020deep,chen2021aminn,lu2021data,lu2021ai}. For example, DeepAttnMISL~\cite{yao2020whole} clustered image patches into different ``bags" to model and aggregate diverse local features for patient-level diagnosis.

However, most prior efforts, especially the ``natural image driven" MIL algorithms, ignore the multi-scale (i.e., pyramidal) nature of WSIs. For example, a WSI consists of a hierarchical scales of images (from 40$\times$ to 5$\times$), which allows pathologists to examine both local~\cite{abousamra2021multi} and global~\cite{abduljabbar2020geospatial} morphological features~\cite{bejnordi2015multi,gao2016multi,tokunaga2019adaptive}. More recent efforts have mimicked such human pathological assessments by using multi-scale images in a WSI~\cite{Hashimoto_2020_CVPR,Li_2021_CVPR}. These methods typically perform independent feature extraction at each scale and then perform a ``late fusion". In this study, we consider the feasibility of examining the interaction between different scales at an earlier stage through an attention-based ``early fusion" paradigm.

In this paper, we propose the addition of a novel cross-scale attention mechanism in an attention-guided MIL scheme to explicitly model inter-scale interactions during feature extraction (Fig.~\ref{fig1:Problem}). In summary, the proposed method not only utilizes the morphological features at different scales (with different fields of view), but also learns their inter-scale interactions as a ``early fusion" learning paradigm. Through empirical validation, our approach achieved the higher Area under the Curve (AUC) scores, Average Precision (AP) scores, and classification accuracy. The contribution of this paper is two-fold:

$\bullet$ A novel cross-scale attention mechanism is proposed to integrate the multi-scale information and the inter-scale relationships.

$\bullet$ Differential cross-scale attention visualizations are generated for lesion pattern guidance and exploration.


\section{Methods}
\begin{figure}[t]
\begin{center}
\includegraphics[width=0.9\textwidth]{{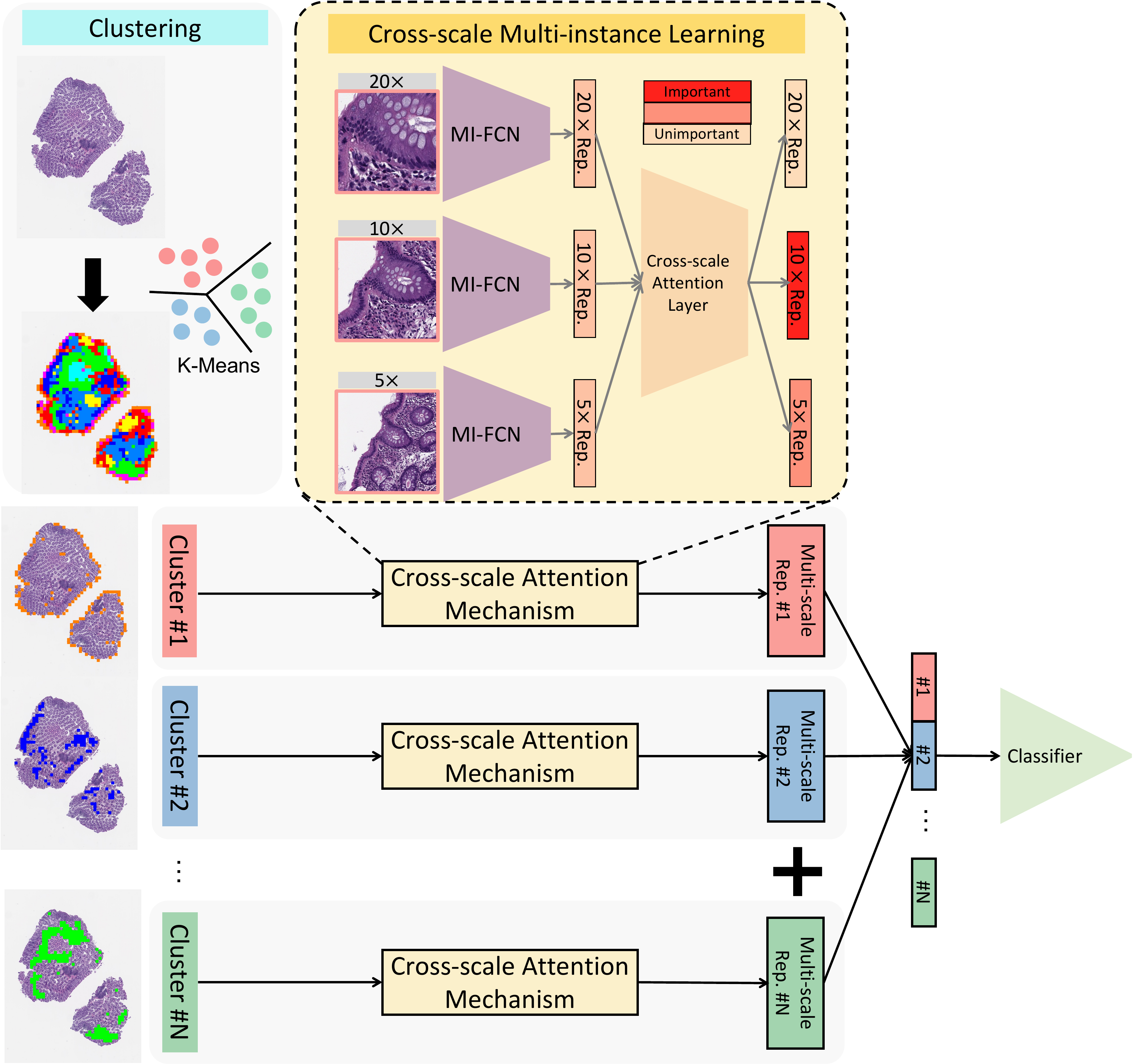}}
\end{center}
\caption{\textbf{Cross-scale Attention Guided Multi-instance Learning Pipeline.} This figure demonstrates the pipeline of the proposed method. The local feature-based clustering was deployed on each WSI to distribute the phenotype patterns in each MIL bag. The cross-scale attention mechanism is deployed in each cluster of MIL branch to combine the multi-scale features with differential attention scores. Multi-scale representations from different clusters were concatenated for CD classification.}
\label{fig2:Pipeline}
\end{figure}

The overall pipeline of the proposed CS-MIL is presented in Fig.~\ref{fig2:Pipeline}. Patches at each location (same center coordinates) at different scales are jointly tiled from WSIs. Patch-wise phenotype features are extracted from a self-supervised model. Then, local feature-based clustering is deployed on each WSI to distribute the phenotype patterns in each MIL bag. Cross-scale attention-guided MIL is proposed to aggregate features in multi-scale and multi-clustered settings. A cross-scale attention map is generated for human visual examination.

\subsection{Feature embedding and phenotype clustering}


In the MIL community, most histopathological image analysis methods are divided into two stages~\cite{schirris2021deepsmile,dehaene2020self}: (1) the self-supervised feature embedding stage and (2) the weakly supervised feature-based learning stage. We follow a similar design that leverages our dataset to train a contrastive-learning model SimSiam~\cite{chen2021exploring} to extract high-level phenotype features from patches. All of the patches are then embedded into low-dimensional feature vectors for the classification in the second stage.

Inspired by~\cite{yao2020whole}, we implement K-means clustering to cluster patches on the patient level based on their self-supervised embeddings from the first stage since the high-level features are more comprehensive than low-resolution thumbnail images in representing phenotypes~\cite{zhu2017wsisa}. When gathering the patches equally from different clusters, the bag with the better generalization for the MIL model can be organized with distinctive phenotype patterns sparsely distributed on WSIs. In contrast, patches with similar high-level features can be aggregated for classification without spatial limitation.


\subsection{Cross-scale attention mechanism}
We implement the MI-FCN encoder from DeepAttnMISL~\cite{yao2020whole} as the backbone to encode patch embeddings from corresponding phenotype clusters and aggregate the instance-wise features to the patient-wise classification, which showed superior performance on survival prediction on WSIs. In the MIL community, several attention mechanisms~\cite{ilse2018attention, lu2021data} have been proposed for instance-relationship between different locations on WSIs. However, those methods are not aware of modeling multi-scale patterns from the pyramid-structured WSIs. Some approaches~\cite{Hashimoto_2020_CVPR,Li_2021_CVPR} have aggregated multi-scale features into deep learning models from WSIs. Unfortunately, those methods fail to exploit relationships between multiple resolutions at the same location.

To address this issue, we propose a cross-scale attention mechanism to represent distinctive awareness at different scales in the backbone. After separately encoding embedding features at different scales, the cross-scale attention mechanism from those encoding features is leveraged to consider the importance of each scale when aggregating multi-scale features at the same location. These attention scores are multiplied by representations from multiple scales to fuse the cross-scale embedding. The multi-scale representation $F$ can be calculated by:

\begin{equation}
    F = \sum_{s=1}^S a_sf_s
\label{eq:cross-scale1}
\end{equation}

\noindent where
\begin{equation}
    a_s = \frac{\exp{\mathbf{W}^\mathrm{T}tanh(\mathbf{V}f_s^\mathrm{T})}}{\sum_{s=1}^S\exp{\mathbf{W}^\mathrm{T}tanh(\mathbf{V}f_s^\mathrm{T})}}
\label{eq:cross-scale2}
\end{equation}

\noindent$\mathbf{W} \in\mathbb{R}^{L \times 1}$ and $\mathbf{V} \in\mathbb{R}^{L \times M}$ are trainable parameters in the cross-scale attention layer. $L$ is the size of the MI-FCN output $f_s$, $M$ is the output channel of the hidden layer in the cross-scale attention layer. Tangent element-wise non-linearity activation function $tanh(.)$ is implemented both negative and positive values for proper gradient flow. $S$ is the number of the scales on WSIs. The attention-based instance-level pooling operator from ~\cite{yao2020whole} is then deployed to achieve patient-wise classification with cross-scale embedding.

\subsection{Cross-scale attention visualisation}

The cross-scale attention maps from the cross-scale attention mechanism on WSIs are presented to show the distinctive contribution of phenotype features at different scales. The cross-scale attentions are mapped from patch scores of the cross-scale attention mechanism on WSIs, demonstrating the importance at multiple resolutions. This attention maps concatenate scale knowledge and location information can expand clinical clues for disease-guiding and exploration in different contexts.



\section{Experiments}
\subsection{Data}
50 H\&E-stained Ascending Colon (AC) biopsies from ~\cite{bao2021cross}, which are representative in CD, were collected from 20 CD patients and 30 healthy controls for training. The stained tissues were scanned at 20$\times$ magnification. For the pathological diagnosis, the 20 slides from CD patients were scored as normal, quiescent, mild, moderate, or severe. The remaining tissue slides from healthy controls were scored as normal. 116 AC biopsies were stained and scanned for testing with the same procedure as the above training set. The biopsies were acquired from 72 CD patients who have no overlap with the patients in the training data.

\subsection{Experimental setting}

256$\times$256 pixels patches were tiled at three scales (20$\times$, 10$\times$ and 5$\times$). For 20$\times$ patches, each pixel is equal to 0.5 Micron. Three individual models following the official SimSiam with a ResNet-50 backbone were trained at three scales, respectively. All three models were trained in 200 epochs with a batch size of 128 with the official setting. 2048-channel embedding vectors were received for all patches. K-means clustering with a class number of 8 was implemented to receive phenotype clustering within the single-scale features at three resolutions, and multi-scale features that include all resolutions for each patient.

10 data splits were randomly organized following the leave-one-out strategy in the training dataset, while the testing dataset was separated into 10 splits with a balanced class distribution. Each bag for MIL models was collected for each patient, equally selecting from different phenotype clustering classes, marked with a slide-wise label from clinicians. Negative Log-Likelihood Loss (NLLLoss)~\cite{yao2019negative} was used to compare the slide-wise prediction for the bag with the weakly label. The validation loss was used to select the optimal model on each data split, while the mean value of the performance on 10 data splits was evaluated as the testing results. Receiver Operating Characteristic (ROC) curves with Area under the Curve (AUC) scores, Precision-Recall (PR) curves with Average Precision (AP) scores, and classification accuracy were used to estimate the performance of each model. We followed the previous work~\cite{gao2021cancer} to implement the bootstrapped two-tailed test and the DeLong test to compare the performance between the different models. The cross-scale attention scores were normalized within every single scale between 0 to 1.

\section{Results}
\subsection{Performance on classification}
We implemented multiple DeepAttnMISL ~\cite{yao2020whole} models with patches at different scales with a single-scale setting. At the same time, we trained the Gated Attention (GA) model ~\cite{ilse2018attention} and DeepAttnMISL model with multi-scale patches, without differentiating scale information. Patches from multiple scales are treated as instances when processing phenotype clustering and patch selection for MIL bags. Furthermore, we adopted a multi-scale feature aggregations, jointly adding embedding features from the same location at different scales into each MIL bag as ~\cite{Hashimoto_2020_CVPR}. We also concatenated embedding features from the same location at different scales as ~\cite{Li_2021_CVPR}. We followed above multi-scale aggregation to input phenotype features into the DeepAttnMISL backbone to evaluate the baseline multi-scale MIL models as well as our proposed method. All of the models were trained and validated within the same hyper-parameter setting and data splits.

\begin{table*}[t]
\caption{Classification Performance on testing dataset.}
\centering
\begin{tabular}{lcccccc}
\hline
Model & Patch Scale & Clustering Scale & AUC & AP & Acc.\\
\hline
DeepAttnMISL(20$\times$)~\cite{yao2020whole} & Single & 20$\times$ &  0.7961 & 0.6764 & 0.7156\\
DeepAttnMISL(10$\times$)~\cite{yao2020whole} & Single & 10$\times$ & 0.7992 & 0.7426 & 0.6897\\
DeepAttnMISL(5$\times$)~\cite{yao2020whole} & Single & 5$\times$ & 0.8390 & 0.7481 & 0.7156\\
\hline
Gated Attention~\cite{ilse2018attention} & Multiple & Multiple & 0.8479 & 0.7857 & 0.7500\\
DeepAttnMISL~\cite{yao2020whole} & Multiple & Multiple & 0.8340 & 0.7701 & 0.7069\\
MDMIL-CNN~\cite{Hashimoto_2020_CVPR} & Multiple & 5$\times$ & 0.8813 & 0.8584 & 0.7759\\
DSMIL~\cite{Li_2021_CVPR} & Multiple & 5$\times$ & 0.8759 & 0.8440 & 0.7672\\
CS-MIL(Ours) & Multiple & 5$\times$ & \textbf{0.8924} & \textbf{0.8724} & \textbf{0.8017}\\
\hline
\end{tabular}
\label{tab:ExtendedDataset}
\end{table*}

\begin{figure}
\begin{center}
\includegraphics[width=0.9\textwidth]{{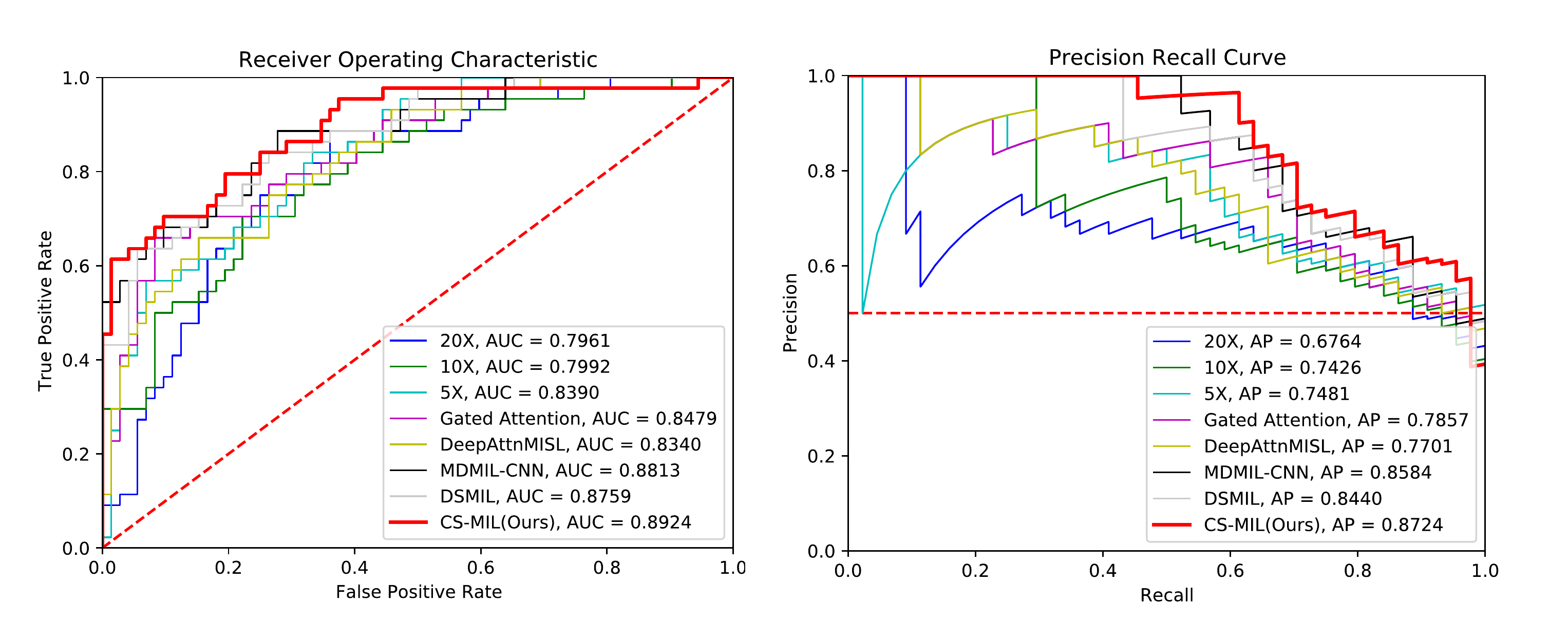}}
\end{center}
\caption{\textbf{ROC curves with AUC scores and PR curves with AP scores.} This figure shows the ROC curves and PR curves of baseline models as well as the AUC scores and AP scores. The proposed model with cross-scale attention mechanism achieved superior performance in two metrics.} 
\label{fig3:Results}
\end{figure}

\begin{figure}[t]
\begin{center}
\includegraphics[width=0.9\textwidth]{{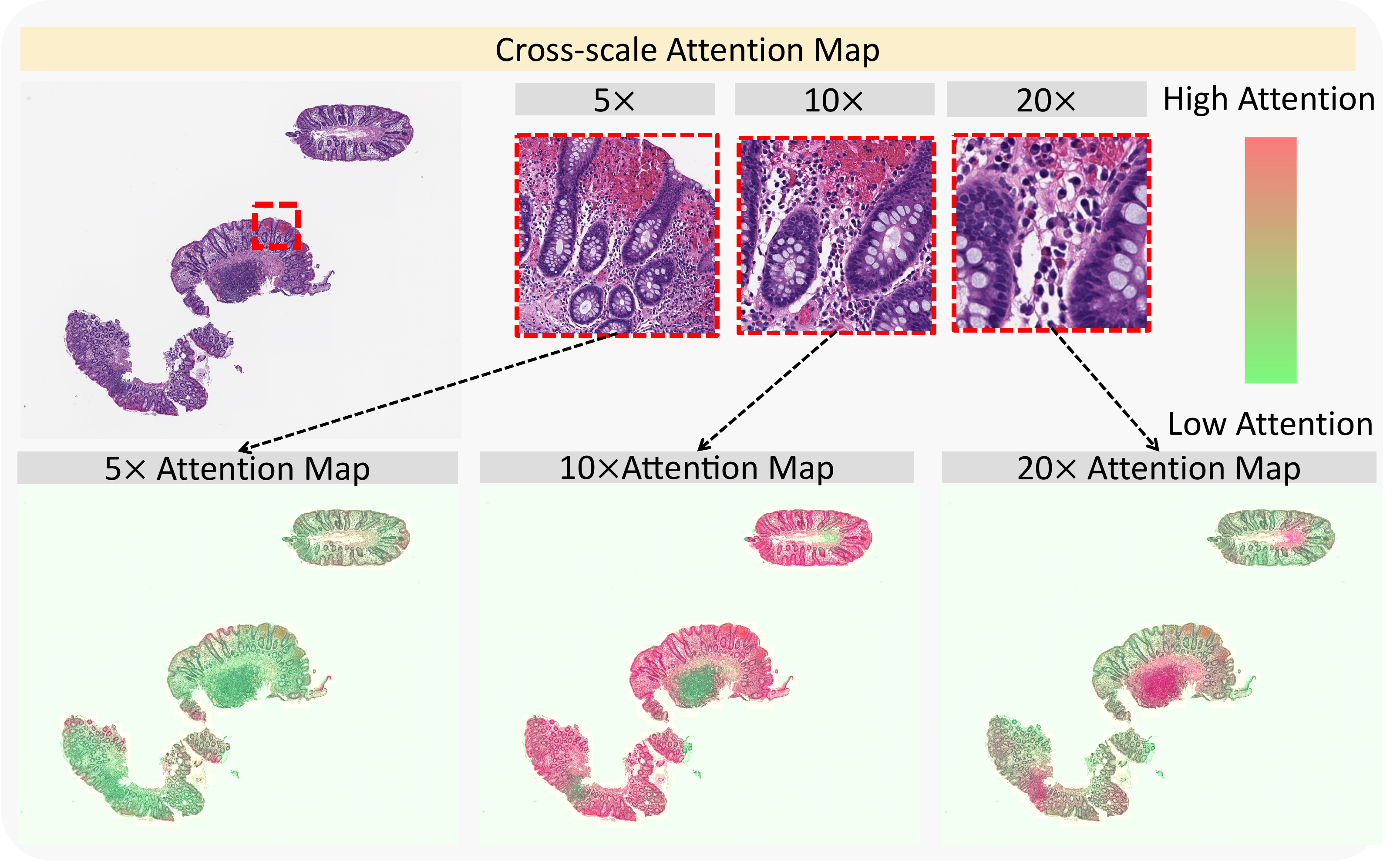}}
\end{center}
\caption{\textbf{Attention Map Visualization.} This figure shows the cross-scale attention maps from the proposed model. The proposed CS-MIL can present importance-of-regions at different scales.} 
\label{fig4:AttentionMap}
\end{figure}

\begin{table*}[t]
\caption{The bootstrapped two-tailed test and the DeLong test between different methods.}
\centering
\begin{tabular}{lcc}
\hline
Model & $p$-value of AUC & $p$-value of AP \\
\hline
DeepAttnMISL(20$\times$)~\cite{yao2020whole} & 0.004 & 0.001\\
DeepAttnMISL(10$\times$)~\cite{yao2020whole} & 0.001 & 0.002\\
DeepAttnMISL(5$\times$)~\cite{yao2020whole} & 0.048 & 0.004\\
\hline
Gated Attention~\cite{ilse2018attention} & 0.070 & 0.031\\
DeepAttnMISL~\cite{yao2020whole} & 0.009 & 0.002\\
MDMIL-CNN~\cite{Hashimoto_2020_CVPR} & 0.466 & 0.457\\
DSMIL~\cite{Li_2021_CVPR} & 0.350 & 0.201\\
CS-MIL(Ours) & Ref. & Ref.\\
\hline
\end{tabular}
\label{tab:Delong}
\end{table*}

\begin{table*}[h]
\centering
\scriptsize
\setlength{\tabcolsep}{2mm}
\renewcommand\arraystretch{1}
\caption{Comparison of different cross-scale attention mechanism designs on testing dataset.}
\begin{tabular}{l|cc|cc|c}
\toprule
Id & Attention Layer Kernel & Activation Function & AUC & AP & Mean of Scores\\
\midrule
1 & Non-sharing & ReLU & 0.8575 & 0.8559 & 0.8576  \\
2 & Non-sharing & Tanh & 0.8848 & 0.8679 & 0.8763  \\
3* & Sharing & ReLU & \textbf{0.8924} & \textbf{0.8724} & \textbf{0.8824}  \\
4 & Sharing & Tanh & 0.8838 & 0.8609 & 0.8723  \\
\bottomrule
\end{tabular}
\label{tab:ablation1}
\end{table*}

\begin{table*}
\centering
\scriptsize
\setlength{\tabcolsep}{2mm}
\renewcommand\arraystretch{1}
\caption{Comparison of different bag sizes on testing dataset.}
\begin{tabular}{l|cc|c}
\toprule
Bag Size & AUC & AP & Mean of Scores\\
\midrule
64 & 0.8507 & 0.8220 & 0.8363  \\
16 & 0.8690 & 0.8523 & 0.8606  \\
08* & \textbf{0.8924} & \textbf{0.8724} & \textbf{0.8824}  \\
01 & 0.8769 & 0.8261 & 0.8515  \\
\bottomrule
\end{tabular}

\label{tab:ablation2}
\end{table*}

\subsubsection{Testing Result}

Table~\ref{tab:ExtendedDataset} and Fig.~\ref{fig3:Results} indicates the performance of the performance while directly applying the models on the testing dataset in the CD classification task, without retraining. In general, single-scale models achieved worse performance compared to multi-scale models, indicating the benefit of external knowledge from multiple scale information. The proposed CS-MIL achieved better scores in all evaluation metrics, showing the benefits of the cross-scale attention which explores the inter-scale relationship at different scales in MIL. Table~\ref{tab:Delong} shows the bootstrapped two-tailed test and the DeLong test to compare the performance between the different models.

\subsubsection{Cross-scale attention visualisation}
Fig.~\ref{fig4:AttentionMap} represents cross-scale attention maps from the cross-scale attention mechanism on a CD WSI and normal WSI. The proposed CS-MIL can present distinctive importance-of-regions on WSIs at different scales, merging multi-scale and multi-region visualization. As a result, the 20$\times$ attention map highlights the chronic inflammatory infiltrates, while the 10$\times$ attention map focuses on the crypt structures. Those regions of interest interpret the discriminative regions for CD diagnosis across multiple scales.

\subsection{Ablation Studies}
Inspired by ~\cite{yao2020whole} and ~\cite{ilse2018attention}, we estimated several attention mechanism designs in MIL with different activation functions. We formed the cross-scale attention learning into two strategies, differentiated by whether they shared the kernel weights while learning the embedding features from multiple scales. We also evaluated the performance of different bag sizes. As a result, as shown in Table~\ref{tab:ablation1},  sharing the kernel weight for cross-scale attention learning with ReLU~\cite{agarap2018deep} achieved better performances with a higher mean value of multiple metrics. Table~\ref{tab:ablation2} demonstrates that a bag size of 8 is an optimal hyper-parameter for this study. The * is the proposed design.

\section{Conclusion}

In this work, we propose the addition of a cross-scale attention mechanism to an attention-guided MIL to combine multi-scale features with inter-scale knowledge. The inter-scale relationship provides extra knowledge of tissues-of-interest in lesions for clinical examination on WSIs to improve the CD diagnosis performance. The cross-scale attention visualization represents automatic scale-awareness and distinctive contributions to disease diagnosis in MIL when learning the phenotype features at different scales in different regions, offering an external AI-based clue for multi-scale pathological image analysis.


\section{Acknowledgements}
This work is supported by Leona M. and Harry B. Helmsley Charitable Trust grant G-1903-03793, NSF CAREER 1452485, and Veterans Affairs Merit Review grants I01BX004366 and I01CX002171, and R01DK103831.




%
%
\bibliographystyle{splncs04}
\bibliography{main}
%




\end{document}